\documentclass[letterpaper, 10 pt, journal, twoside]{ieeetran}
\usepackage{times}
\usepackage[pdftex]{graphicx}
\usepackage{}
\usepackage{amsmath,amssymb,amsopn,amstext,amsfonts}
\usepackage{pifont}

\usepackage{cancel}
\usepackage[space]{cite}
\usepackage{pdfsync}
\usepackage{balance}
\usepackage{color}
\usepackage{mathtools}

\usepackage{setspace}

\usepackage[ruled,vlined]{algorithm2e}

\usepackage{booktabs}
\usepackage{makecell}
\usepackage{bm}

\usepackage{diagbox}
\usepackage{epstopdf}
\usepackage{pifont}
\usepackage{amsmath}
\usepackage{multirow}
\usepackage{url}
\usepackage{verbatim}
\usepackage{footmisc}
\usepackage[linkcolor=black,citecolor=black,urlcolor=black,colorlinks=true]{hyperref}
\usepackage{color}
\usepackage{tabularx}
\usepackage{float}
\usepackage{caption}
\usepackage{geometry}
\graphicspath{{figures/}}
\DeclareGraphicsExtensions{.png,.jpg,.eps,.pdf}

\title{MARSIM: A light-weight point-realistic simulator for LiDAR-based UAVs}
\author{
Fanze Kong$^{1}$, Xiyuan Liu$^{1}$, Benxu Tang$^{2}$, Jiarong Lin$^{1}$, Yunfan Ren$^{1}$,

Yixi Cai$^{1}$, Fangcheng Zhu$^{1}$, Nan Chen$^{1}$, Fu Zhang$^{1}$
\thanks{
$^{1}$These authors are with the Department of Mechanical Engineering, University of Hong Kong. {\tt\small \{kongfz,xliuaa,jiarong.lin,
renyf,yixicai,zhufc,cnchen\}@connect.hku.hk and \{fuzhang\}@hku.hk}. $^{2}$Benxu Tang is with the School of Mechanical Engineering and Automation, Harbin Institute of Technology. {\tt\small 180320222@stu.hit.edu.cn.} (Corresponding author: \textit{Fu Zhang}).}
}

\begin{document}

\maketitle



\begin{abstract}

The emergence of low-cost, small form factor and light-weight solid-state LiDAR sensors have brought new opportunities for autonomous unmanned aerial vehicles (UAVs) by advancing navigation safety and computation efficiency. Yet the successful developments of LiDAR-based UAVs must rely on extensive simulations.
Existing simulators can hardly perform simulations of real-world environments due to the requirements of dense mesh maps that are difficult to obtain. In this paper, we develop a point-realistic simulator of real-world scenes for LiDAR-based UAVs. The key idea is the underlying point rendering method, where we construct a depth image directly from the point cloud map and interpolate it to obtain realistic LiDAR point measurements.
Our developed simulator is able to run on a light-weight computing platform and supports the simulation of LiDARs with different resolution and scanning patterns (e.g., spinning LiDARs and solid-state LiDARs), dynamic obstacles, and multi-UAV systems. Developed in the ROS framework, the simulator can easily communicate with other key modules of an autonomous robot, such as perception, state estimation, planning, and control. Finally, 
{the simulator provides 10 high-resolution point cloud maps of various real-world environments, including forests of different densities, historic building, office, parking garage, and various complex indoor environments. These realistic maps provide diverse testing scenarios for an autonomous UAV. Evaluation results show that the developed simulator achieves superior performance in terms of time and memory consumption against Gazebo and that the simulated UAV flights highly match the actual one in real-world environments. We believe such a point-realistic and light-weight simulator is crucial to bridge the gap between UAV simulation and experiments and will significantly facilitate the research of LiDAR-based autonomous UAVs in the future.}



\end{abstract}



\section{Introduction}


Recent developments of LiDAR technologies have significantly lowered the cost and weight of LiDAR sensors, which creates many opportunities for unmanned aerial vehicle (UAV) applications, such as mine exploration\cite{dang_graphbased_2020}, biological data statistics\cite{shah2020multidrone}, mapping\cite{vacanas2015building}, high-speed navigation \cite{bubbleplanner}, and obstacle avoidance, etc. However, deploying UAVs to these widespread applications requires extensive tests, which are often cost-demanding since the system under test are still in active development and hence may have a noticeable failure rate  (e.g., collision with the environment). A simulator that resembles the reality can significantly reduce the time and equipment cost occurred in UAV tests and has become a crucial component of UAV developments. 

\begin{figure}[t]
	\centering
	\includegraphics[width=0.49\textwidth]{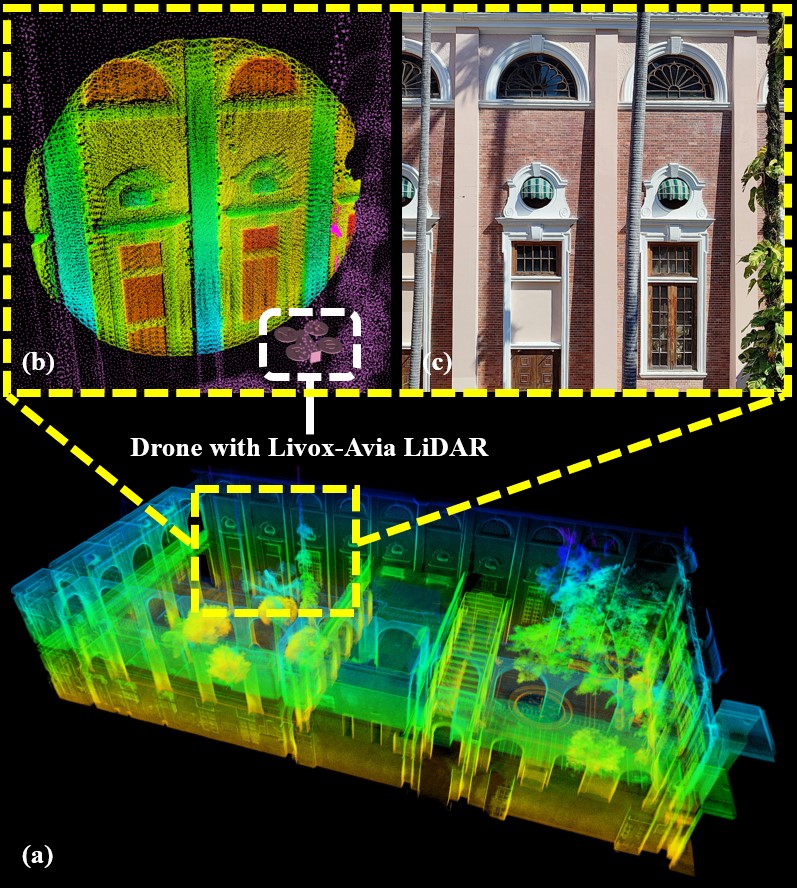}
	\caption{A demo of MARSIM. (a) is the point cloud map of the HKU main building, one of ten real-world scenes of MARSIM; (b) A scan of points of a Livox Avia LiDAR rendered directly from the point cloud map by our simulator, and (c) is the photo of the corresponding scene in real world. It can be seen that the simulator can restore the structural details of the real scene with a high quality. More details can be found in the attached video (also available at \url{https://youtu.be/hiRtcq-5lN0}). }
	\label{fig:coverfigure}
 \vspace{-10mm}
\end{figure}

Most existing simulators (e.g., Gazebo\cite{gazebo}, Webots\cite{webots}, Airsim\cite{shah2018airsim}) have difficulties meeting the demand of high-resolution realistic scene simulation for LiDAR-based UAVs due to the following limitations: (i) the environments that the mainstream simulators can simulate are mostly virtual, unrealistically simple, man-made environments, which possess a considerable gap from complex real-world scenes; 
(ii) existing simulators only import mesh maps, which are difficult to obtain from real-world environments that are often measured in the form of 3D point cloud data by laser scanners or LiDARs. To the best of our knowledge, there are no open-source and mature tools available for generating high-resolution and high-fidelity mesh maps out of point cloud data. The commonly-used Poisson reconstruction \cite{poisson} method is time-consuming and has low-quality meshes on real point cloud data captured by LiDARs due to occlusions and point density variations in large scene scanning; (iii) the mainstream simulators often rely on high-performance GPUs to achieve real-time simulations in large complex mesh maps, which puts a high requirement for computing platforms.

Motivated by these gaps, in this paper, we propose a light-weight LiDAR-based UAV simulator, which has the following features:

\begin{enumerate}

\item Directly utilizing point cloud maps of real environments to render realistic LiDAR scans. The point cloud map keeps fine details of the environments and can be easily obtained using a LiDAR sensor.
\item Highly efficient in computation and memory consumption, and able to run on personal computers without a dedicated graphics processing unit (GPU) board. 
\item Versatile, supporting the simulation of dynamic obstacles, multi-UAV systems, and various existing LiDAR models of different resolutions and scanning patterns (Livox AVIA, Livox MID-360, Velodyne VLP-32, IntelRealsense D455, etc.).  
\item Open-source and ROS-compatible (\url{https://github.com/hku-mars/MARSIM.git}). Users can easily integrate the simulator with their modules developed in ROS, such as simultaneous localization and mapping (SLAM) and path planning modules, and conduct evaluations in realistic simulation environments swiftly.

\end{enumerate}

\section{Related Works}

\begin{figure*}[t]
    \begin{center} 
    \includegraphics[width=0.99\textwidth]{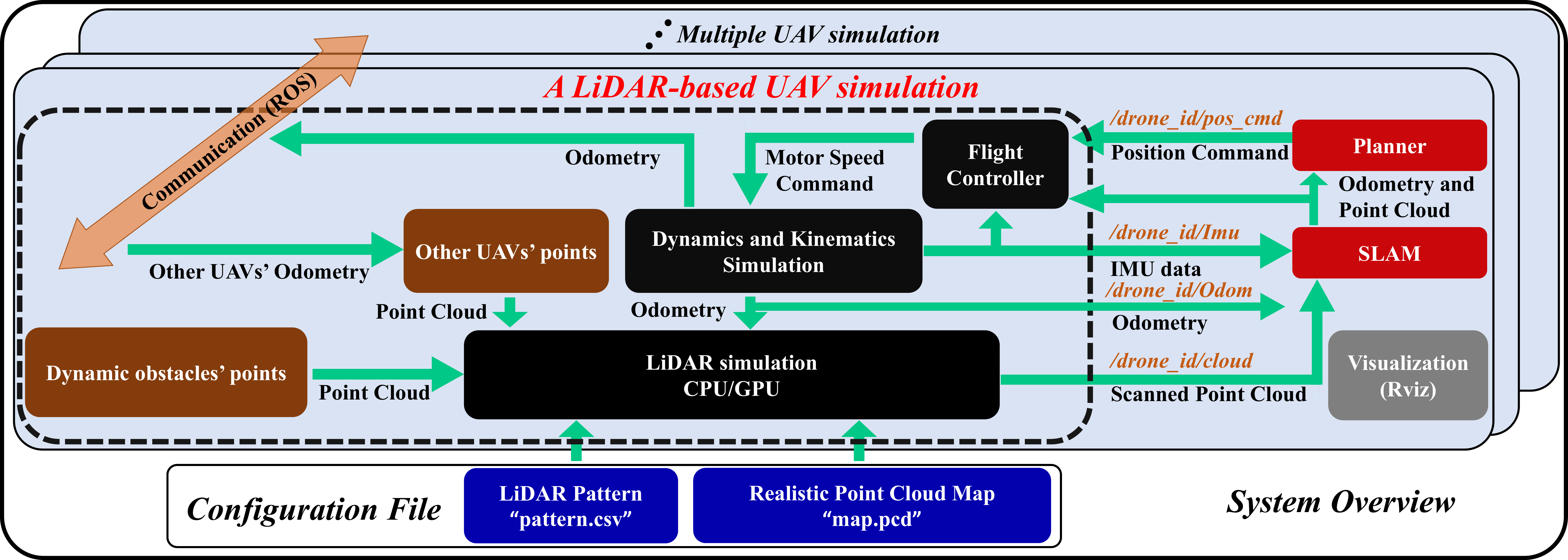}
    \caption{The overall framework of our simulator (black dashed box) and how it interacts with external modules in ROS.}
    \label{fig:simulator_block}
    \end{center}
    \vspace{-5mm}
\end{figure*}

LiDAR-based UAV simulation consists of UAV motion simulation and LiDAR simulation. Compared to the LiDAR simulation, the UAV motion simulation is rather straightforward and mature. MATLAB can support many types of UAVs' motion simulation and controller design, such as quadrotors\cite{matlab_quadrotor} and VTOL UAVs \cite{lvximin_tailsitter}. Common simulators such as Gazebo\cite{gazebo} and Airsim\cite{shah2018airsim} are also able to simulate the movement of UAVs at high frequencies. Wei \textit{et al.} \cite{gazeboplanning} utilize the Gazebo simulation to develop a multi-UAV path planning algorithm, and Han \textit{et al.} \cite{airsim_gaofei} simulate a UAV SE(3) planning in Airsim and provide a benchmark for autonomous UAV racing.

The main challenge of LiDAR-based UAV simulation lies in the LiDAR simulation. There are many existing simulators that support LiDAR simulations, such as Gazebo\cite{gazebo}, Webots\cite{webots}, Airsim\cite{shah2018airsim} and, SVL\cite{rong_lgsvl_2020}. Gazebo is probably the most commonly used simulation platform for mobile robot research, where users can build their own robots and LiDAR sensors. This simulator has been widely used for the verification of autonomous LiDAR-based exploration algorithms, such as GBP\cite{dang_graphbased_2020}, MBP\cite{mbplanner}, TARE\cite{cao_tare_2021} and Splatplanner\cite{splatplanner}. SVL simulates a LiDAR mounted on a vehicle rooftop in an urban environment for the application of autonomous driving. 

The main drawback of these simulators is that they import maps in the form of mesh models, which are often available for artificial environments modeled by 3D modeling software (e.g., Sketchup, Blender, 3DS Max, etc.) or built based on Gazebo model library. For instance, Splatplanner made eleven artificial maps in Gazebo\cite{splatplanner,brunel_flybo_2021}, and TARE made five larger artificial maps in Gazebo for algorithm verification. Besides, DARPA subterranean competition\cite{darpa} provides a set of tunnel and mine maps where most of them are manual-made. Artificial, manual-made maps can meet some basic simulation requirements, but most of them are relatively simple and unrealistic, leading to a large gap from complex real-world environments. 


A more common representation form of real-world environments is point cloud, which can be collected by devices such as 3D laser scanners or LiDAR sensors. To fill the gap between simulation and reality, some existing works attempt to provide realistic mesh maps from point clouds. For example, LiDARsim\cite{Lidarsim} generates realistic mesh maps using surfel-based method\cite{surfels} for realistic self-driving simulation. Other methods have also been developed to construct realistic mesh maps from point clouds, such as Poisson reconstruction\cite{poisson}, Truncated Signed Distance Field (TSDF) volume method\cite{tsdf}, and Delaunay triangulation method\cite{delauncytriangulation,r3live}. However, these methods are not robust to occlusion and point density variation that often occur in point clouds collected by LiDARs in large-scale scenes. In this case, non-existent surfaces may be falsely generated which require extra labor work to fix or tune the parameter. To overcome these issues, FlightGoggles\cite{flightgoggles} separately construct each individual object in the scene in a commercial software \textit{Reality Capture} and then synthesize them with the background to build a highly realistic simulation environment. This approach is not very scalable since is very time- and labor-consuming to model each object and synthesize the scene (FlightGoggles\cite{flightgoggles} provides only two scenes).  Finally, high-resolution mesh models are also challenging to render in real time without high-performance GPUs. 
{A question arises whether we really need mesh maps for LiDAR simulation.}
Mesh models have the advantage to attach material texture recovered from camera images, which can then render images for camera simulation. However, for LiDAR simulation, such texturing is not necessary. Moreover, thanks to the recent developments of low-cost, high-precision LiDAR sensors and simultaneous localization and mapping (SLAM) algorithms \cite{fastlio2, xiyuanBA}, obtaining high-precision point clouds of real-world environments are becoming much more affordable and accessible. Motivated by this trend, we choose to directly use point cloud maps for simulation instead of mesh maps. A similar idea has been preliminarily explored in FUEL\cite{zhou_fuel}, but the presented simulator has limited resolution and accuracy that is suitable for only depth cameras in small scenes due to the high computation cost.



\section{System Overview}

As shown in Fig. \ref{fig:simulator_block}, our UAV simulator is mainly composed of three submodules: a built-in flight controller module, a dynamics and kinematics simulation module, and a LiDAR simulation module (modules in black, see Fig. \ref{fig:simulator_block}). The simulator is able to interact with planners, SLAM algorithms, and visualization modules in the ROS framework, forming a complete LiDAR-based UAV simulation system.

To use the simulator, users should first choose a LiDAR model and supply a point cloud map of the environment. Users can then plug in their own SLAM (or use ground-true odometry) and Planner algorithms to the UAV simulator via the ROS topic names shown in Fig. \ref{fig:simulator_block} for verification and visualization. Once the simulator starts, the dynamics and kinematics simulation module starts to compute the UAV's odometry and IMU data, according to which the LiDAR simulation module then renders the LiDAR scanned point cloud. The simulated IMU data and LiDAR scans are published in ROS, which could be used by the SLAM and then by the planner module. Besides the static environments represented by the point cloud map, the LiDAR simulation also simulates point measurements on dynamic obstacles and other UAVs in real time. 

\section{Methodology}
\label{section:software architecture}

\subsection{LiDAR simulation}

Given a point cloud representation of the environment (see Sec. \ref{section:highresolutionmapping}) and the current LiDAR pose, the LiDAR simulation module aims to render the points that should be measured in the current LiDAR scan. To do so, we first project all points of the point cloud map into the current LiDAR FoV and perform interpolation to obtain a dense depth image of 
$\theta_{\text{res}}$ angular 
resolution and then mask those points that are not on the scanning pattern. The remaining depth pixels are finally added with LiDAR measurements noises and transformed to point clouds for publishing (see Fig. \ref{fig:depthimage}). To accelerate the point projection process, we perform two key preprocessing when importing the point cloud map: 1) to limit the number of map points, we conduct spatial downsampling with a spatial resolution $r_{\text{map}}$ on all map points; 2) we cut the map space into equal large voxels (with cube length $l$) and save the map points contained in each voxel in its respective point list. When rendering the LiDAR point cloud, we {screen out voxels having intersection with the current LiDAR FoV, only points in these voxels are projected. }

\subsubsection{Occlusion culling}

\begin{figure}[t]
	\centering
	\includegraphics[width=0.49\textwidth]{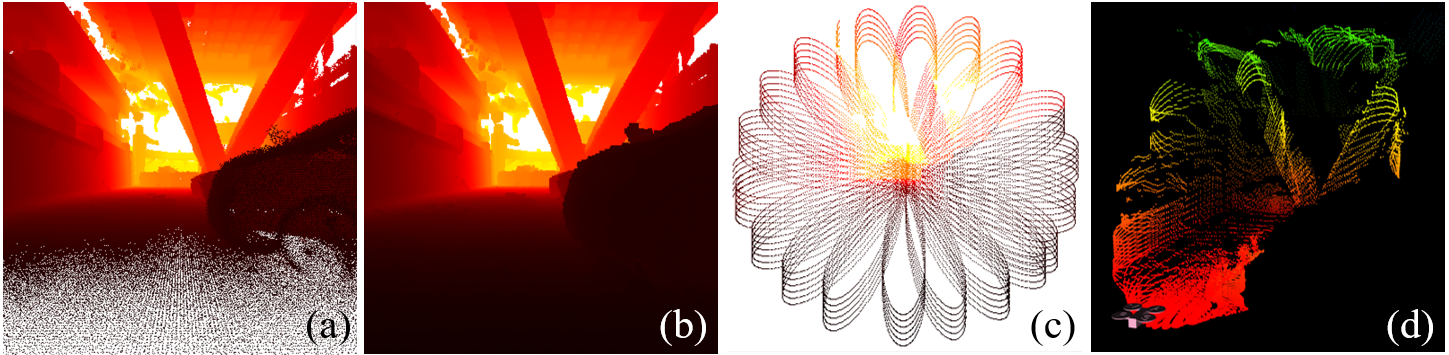}
	\caption{
	 Illustration of LiDAR simulation. (a) is the sparse depth image which directly projects map points to the image. (b) is the dense image after interpolation. (c) shows the depth image masked with the LiDAR scanning pattern (a Livox Avia LiDAR). (d) shows the final output LiDAR point clouds.}
	\label{fig:depthimage}
	\vspace{-10mm}
\end{figure}

While effectively limiting the number of map points, spatial downsampling will cause two problems. One is that the projected points do not uniformly populate the depth image due to the projection principle: points close to the LiDAR sensor are very sparse while those that at far are very dense, resulting in many empty depth pixels (as shown in Fig. \ref{fig:depthimage} (a)). {Another problem is that after projected to the depth image, points on background objects will intervene in points on foreground objects and hence cause false depth measurements.} To address these issues, we perform an occlusion culling process as follows.


Due to the spatial downsampling with resolution $r_{\text{map}}$, a point in the map should really represent a cube of substance with length $r_{\text{map}}$ (see Fig. \ref{fig:interpolation}(a)). Therefore, the interpolation range $\theta_{\text{max}}$ around a map point $\mathbf p$ on the depth image is the projected area of the point's corresponding cube:

\begin{equation}
	\theta_{\text{max}} = {\text{arcsin}} \left(\frac{\frac{\sqrt{3}}{2}r_{\text{map}}}{d} \right)
	\label{eq:1}
\end{equation}
where $d$ is the depth of the map point $\mathbf p$. Pixels within the interpolation range $\theta_{\text{max}}$ will have their depth values set to $d$.
If the interpolation ranges of different map points have overlaps, the overlapped pixels will retain the smallest depth value.

\subsubsection{Plane correction}

\begin{figure}[t]
    \centering
    \includegraphics[width=0.49\textwidth]{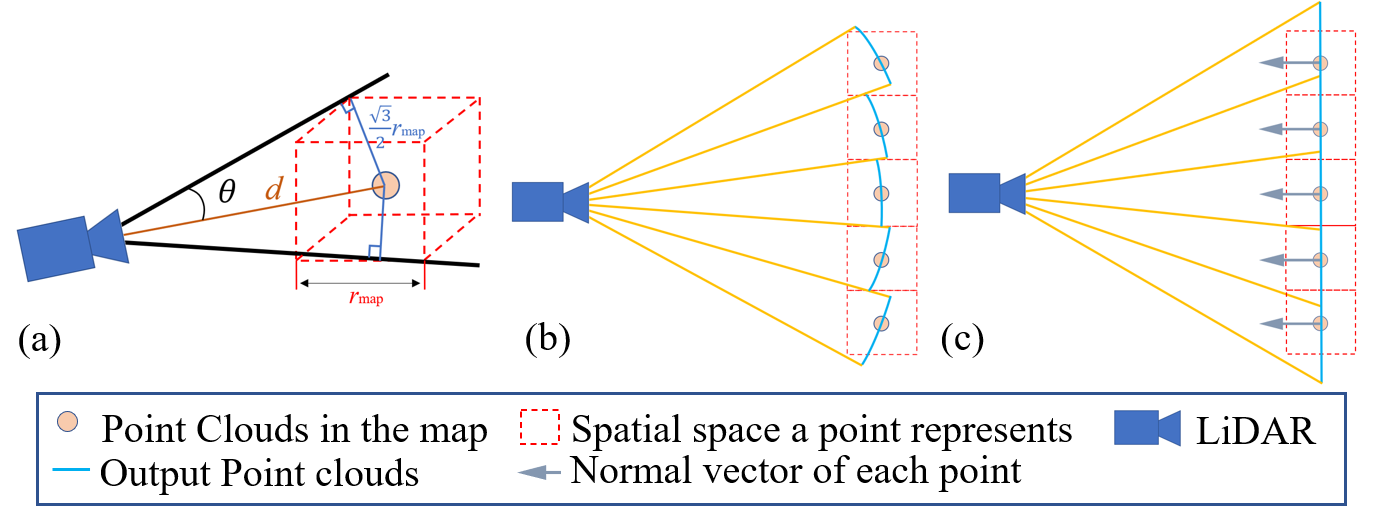}
    \caption{The principle of occlusion culling and plane correction. (a) demonstrates the geometric principle to compute the interpolation range of each point. (b) demonstrates the problem of occlusion culling on large planes and (c) demonstrates the plane correction.}
    \label{fig:interpolation}
    \vspace{-18mm}
\end{figure}

{The occlusion culling works well when the LiDAR laser ray is perpendicular to the object's surface. }
{However, when the laser ray is not perpendicular to the surface, the interpolated points will go out of the surface (see blue point clouds in Fig. \ref{fig:interpolation} (b)), which causes unexpected false point measurements. }
To address this issue, when interpolating the neighbor of a map point, we view the map point as a small plane and 
interpolate pixels in the interpolation range $\theta_{\text{max}}$ by calculating the intersection point between the pixel ray and the plane (see Fig. \ref{fig:interpolation} (c)). The small plane around each map point is fitted from its neighboring points in the map during the preprocessing stage and the estimated plane normal is saved along with the map point. Since the depth calculation from a plane is more time-consuming, we perform such plane correction only for map points genuinely on a plane. This is achieved by introducing a new variable $\gamma$ of each map point, which indicates the plane quality of the point. The value $\gamma$ is computed as the plane thickness during the plane fitting and is saved along with the plane normal and map point. During online rendering, only map points with $\gamma$ below a certain threshold will perform the plane correction.

\subsubsection{GPU acceleration}

When the map resolution increases, the size of point clouds expands gradually to millions, and the time for projecting map points onto the depth image will increase considerably. In this case, only using the central processing unit (CPU) can no longer meet the real-time requirements. To address this issue, we utilize Graphics Processing Unit (GPU) hardware to accelerate the LiDAR simulation process. Affordable CPU-integrated or standalone GPUs are widely available in standard personal computer and hence does not degrade the generality of our simulator. They are also well supported by Open Graphics Library (OpenGL)\cite{opengl}, which is a multi-platform general graphics rendering library that can efficiently use GPU resources to project point clouds onto depth image in our task. We propose a point cloud parallel renderer based on OpenGL to accelerate the depth image construction.
With GPU acceleration, the simulator is able to run in real time (more than  10 Hz) on point cloud maps with tens of millions of points in the map.

\subsection{Dynamics and kinematics simulation}

In order to simulate a realistic UAV flight, the simulator provides a dynamics and kinematics simulation of UAV according to the standard rigid-body model\cite{dynamicmodel}. The thrust and torque in the dynamic model are generated from a second-order motor model whose command is the expected motor speed\cite{propellermodel}. 
The model parameters (e.g., inertia matrix, mass, propeller torque, thrust coefficients, motor KV value) are drawn from\cite{dynamicmodel} and are summarized in a separate configuration file where users may modify as needed. The computed angular velocity and special acceleration are added with measurement noises and biases to obtain the IMU measurements for publishing. The complete ground-true UAV state is also published for external reference. 

\subsection{Flight controller design}

After the motion simulation of a UAV, a controller is needed to accurately control the UAV flight. Our simulator adopts a cascaded dual-loop PID controller as shown in \cite{moderncontrol}, where the inner loop is an attitude controller, and the outer loop is a position controller. According to the UAV dynamic and kinematic model parameters, the controller gains are tuned based on the expected natural frequency and system damping ratio to achieve a good position control performance.


\subsection{Dynamic obstacles simulation and collision detection}

To mimic the real environment where dynamic obstacles may appear, the simulator supports the simulation of dynamic obstacles. We randomly generate a certain number of spherical dynamic objects and point clouds on the objects' surfaces. Each spherical object moves at a constant velocity in a random direction. When one spherical object moves out of the map boundary, a new one is generated at a random location in the map. The number, size, and speed of the spherical objects can be modified by users if necessary. 

To best simulate a reality, the simulator conducts a collision check at each simulation step to detect if the UAV collides with obstacles (static or dynamic) in the environment. To achieve this, the simulator establishes two KD-Trees, one is built from the global map points at the preprocessing stage and the other is built from points on the dynamic objects at each simulation step. Then, the simulator searches for nearest neighbor points within the range equal to the UAV size. If any points are found in the range, the UAV is deemed as colliding with obstacles, and the simulator would output a collision warning.


\begin{figure*}[t!]
	\centering
	\includegraphics[width=0.99\textwidth]{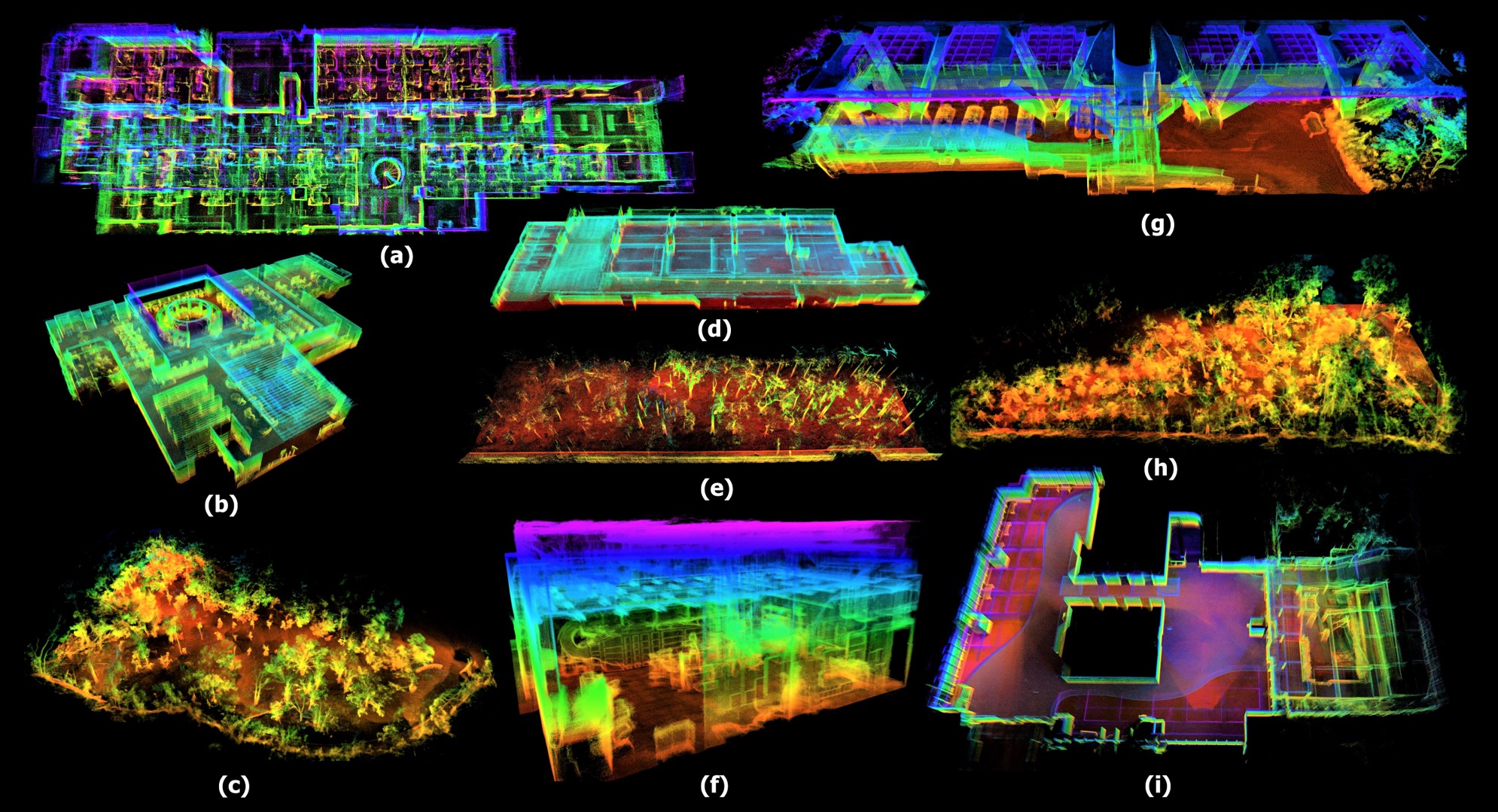}
	\caption{High-resolution point cloud maps provided by the simulator with detailed information shown in Table \ref{tab1:Mapscharateristic}.}
	\label{fig:allmaps}
\end{figure*}



\begin{table*}[h!]
		\centering
	\caption{{Ten realistic point cloud maps and their features.}}		\begin{tabular}{@{}llcccccc@{}}
			\toprule
			\textbf{MARSIM Maps}& Size (m) & \makecell[c]{Large-\\scale} & \makecell[c]{Multi-\\layer} & \makecell[c]{Cluttered \\Obstacles} & \makecell[c]{Narrow \\corridor} &
			\makecell[c]{Thin \\ Structures} & \makecell[c]{Real-world \\ Environments}\tabularnewline
			\midrule
			Historical Building (Fig. \ref{fig:coverfigure} (c)) & 47$\times$20$\times$23 &  & \checkmark & \checkmark & & \checkmark & \checkmark\tabularnewline
			Large Office (Fig. \ref{fig:allmaps} (a)) & 45$\times$16$\times$6 & \checkmark & \checkmark & \checkmark & \checkmark & \checkmark & \checkmark\tabularnewline
			Indoor-2 (Fig. \ref{fig:allmaps} (b)) & 27$\times$40$\times$7 & & & \checkmark & \checkmark &  & \checkmark\tabularnewline
			Common Forest (Fig. \ref{fig:allmaps} (c)) & 48$\times$27$\times$19 & \checkmark & & \checkmark & & \checkmark & \checkmark\tabularnewline
			Simple Parking Garage (Fig. \ref{fig:allmaps} (d)) & 45$\times$15$\times$5 & \checkmark & & & & & \checkmark\tabularnewline
			Simple Forest (Fig. \ref{fig:allmaps} (e))& 45$\times$16$\times$6 & \checkmark & & \checkmark & & \checkmark & \checkmark\tabularnewline
			Indoor-1 (Fig. \ref{fig:allmaps} (f))& 17$\times$13$\times$9 & & & \checkmark & \checkmark & \checkmark & \checkmark\tabularnewline
			Complex Parking Garage (Fig. \ref{fig:allmaps} (g)) & 62$\times$12$\times$10 & \checkmark & & \checkmark & & & \checkmark\tabularnewline
			Dense Forest (Fig. \ref{fig:allmaps} (h)) & 30$\times$72$\times$19 & \checkmark & & \checkmark & \checkmark & \checkmark & \checkmark\tabularnewline
			Indoor-3 (Fig. \ref{fig:allmaps} (i)) & 21$\times$48$\times$4 & & & \checkmark &  & \checkmark & \checkmark\tabularnewline
			\bottomrule
		\end{tabular}
	\label{tab1:Mapscharateristic}
\end{table*}

\subsection{Decentralized Multiple UAV simulation}

A notable trend in UAV research is swarm navigation and control. To enable such research, our simulator supports the simulation of UAV swarm systems. To distribute the computation load and make the simulator more scalable to swarm size, the simulation is completely decentralized, where each UAV is simulated in one separate thread (as a ROS node) or computer. Different threads or computers communicate via ROS communication. This way, multiple UAV simulations can be distributed to multiple light-weight computers with local area network (LAN) connection. In order to simulate the interaction between multiple UAVs more realistically, the simulator adds a mutual observation function: at each simulation step, a set of points are sampled on a UAV surface (approximated as a cube of the UAV size), the sampled points are then added to the global point cloud map to render the rest UAVs' LiDAR scan. 



\section{Realistic high-resolution point cloud map construction}
\label{section:highresolutionmapping}
To enable realistic interaction between the UAV and the environment in the simulator, we construct point cloud maps from actual scenes. In order to restore the realistic environment as much as possible, we put forward two requirements for the point cloud map: high resolution and high precision. To fulfill these requirements, we use a hand-held device carrying a Livox Avia sensor (detailed in\cite{fastlivo}) to scan the environment. The non-repetitive scanning of Livox Avia LiDAR enables the map points to be accumulated to a high resolution even when the LiDAR is placed at a stationary position. This reduces the effect of point density variation caused by LiDAR motion. To register all LiDAR scans under the same global frame, we use FAST-LIO2\cite{fastlio2} to construct a rough map and then utilize \cite{xiyuanBA} to globally refine the map quality by LiDAR bundle adjustment. The globally registered point cloud map is imported to CloudCompare software, where statistical outlier removal (SOR) filter and spatial downsampling are used to generate a uniform clean point cloud map. SOR filter is a filter that computes each point's average distance to its neighbors and removes the points having a relatively large distance (i.e., isolated noisy points). {After filtering the outlier points, we manually fill some points in corners that the LiDAR scan cannot reach in real world.}
Using the above methods, we scanned ten real-world environments and obtained corresponding high-precision point cloud maps for the use of the simulator. The method can also be used to obtain point cloud maps of man-mand environments represented by mesh models if needed. 


\section{Results}
\label{section:results}
\subsection{High-resolution realistic Point Cloud Maps}

\begin{figure}[t!]
	\centering
	\includegraphics[width=0.49\textwidth]{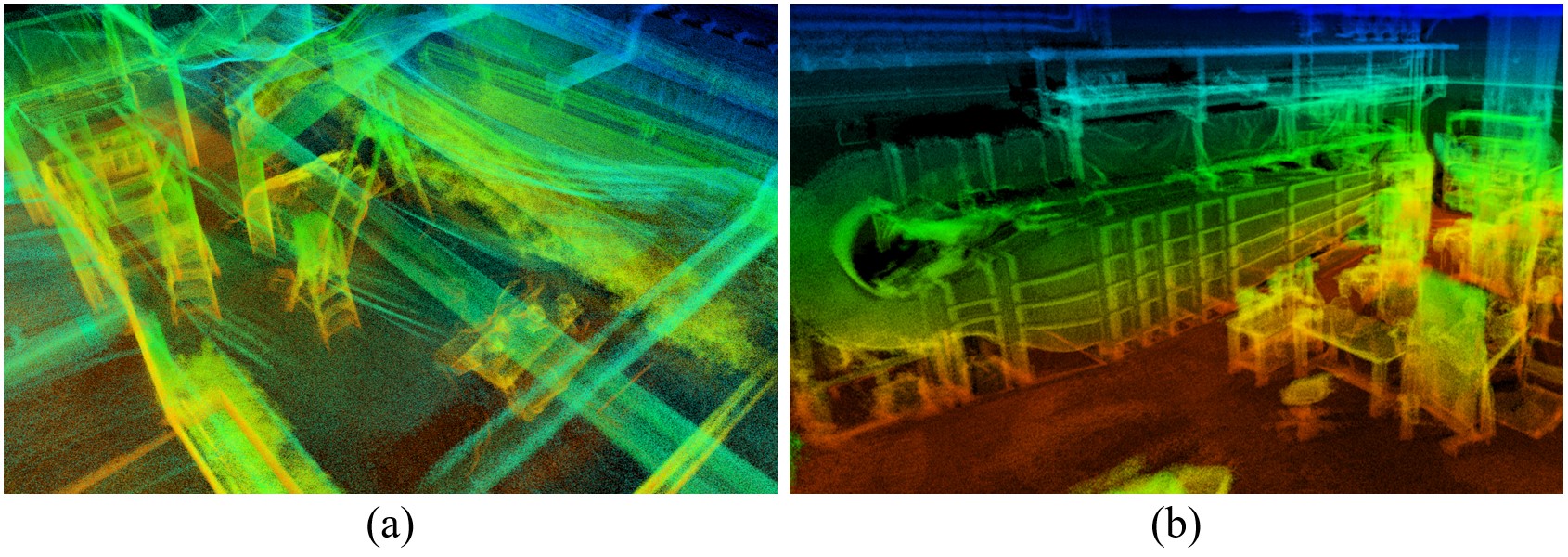}
	\caption{Detailed structure of indoor-1 and indoor-2 maps. (a) shows several clear ladders in indoor-1 map and (b) shows many complex equipment in a machine workshop.}
	\label{detailsofmaps}
\end{figure}

This paper provides high-resolution (0.01 m) point cloud maps of ten real scenes for users to simulate, as shown in Fig. \ref{fig:allmaps}. Some of their environmental features can be seen in Table \ref{tab1:Mapscharateristic}, which can be used for reference when choosing a simulation map. The 0.01-m resolution map here refers to the original point cloud processed by the 0.01-m spatial downsampling. The scenes of the ten maps are three forests, three indoor scenes, a historical building (the HKU main building), two parking garages, and a large office. Detailed structures in real environments can be clearly seen in the close-up point clouds shown in Fig. \ref{detailsofmaps}.


\begin{figure}[t]
	\centering
	\includegraphics[width=0.49\textwidth]{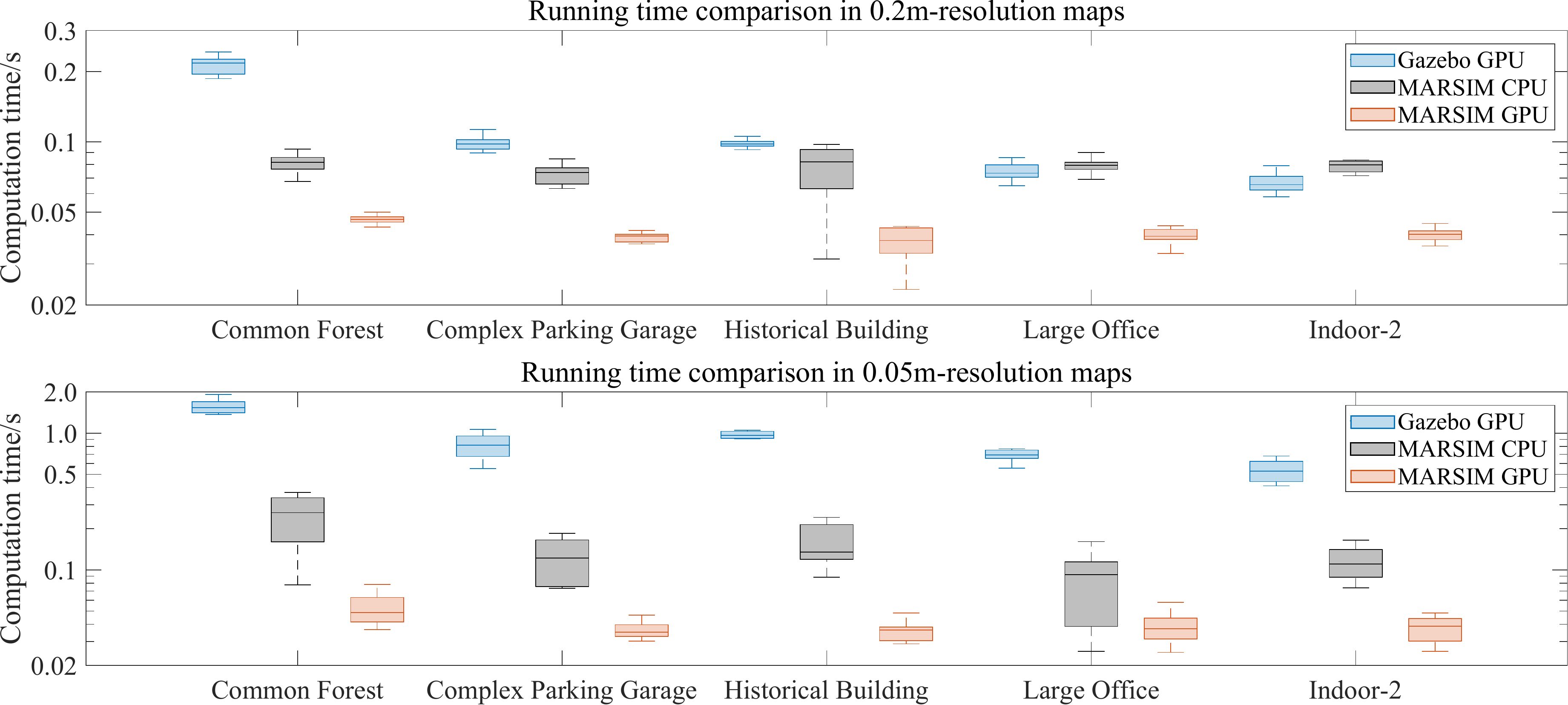}
	\caption{{Time consumption for rendering one Livox AVIA scan on a light-weight computation platform (NUC). }}
	\label{Gazebocomparison}
 \vspace{-15mm}
\end{figure}

	\begin{figure}[t]
		\centering
		\includegraphics[width=0.49\textwidth]{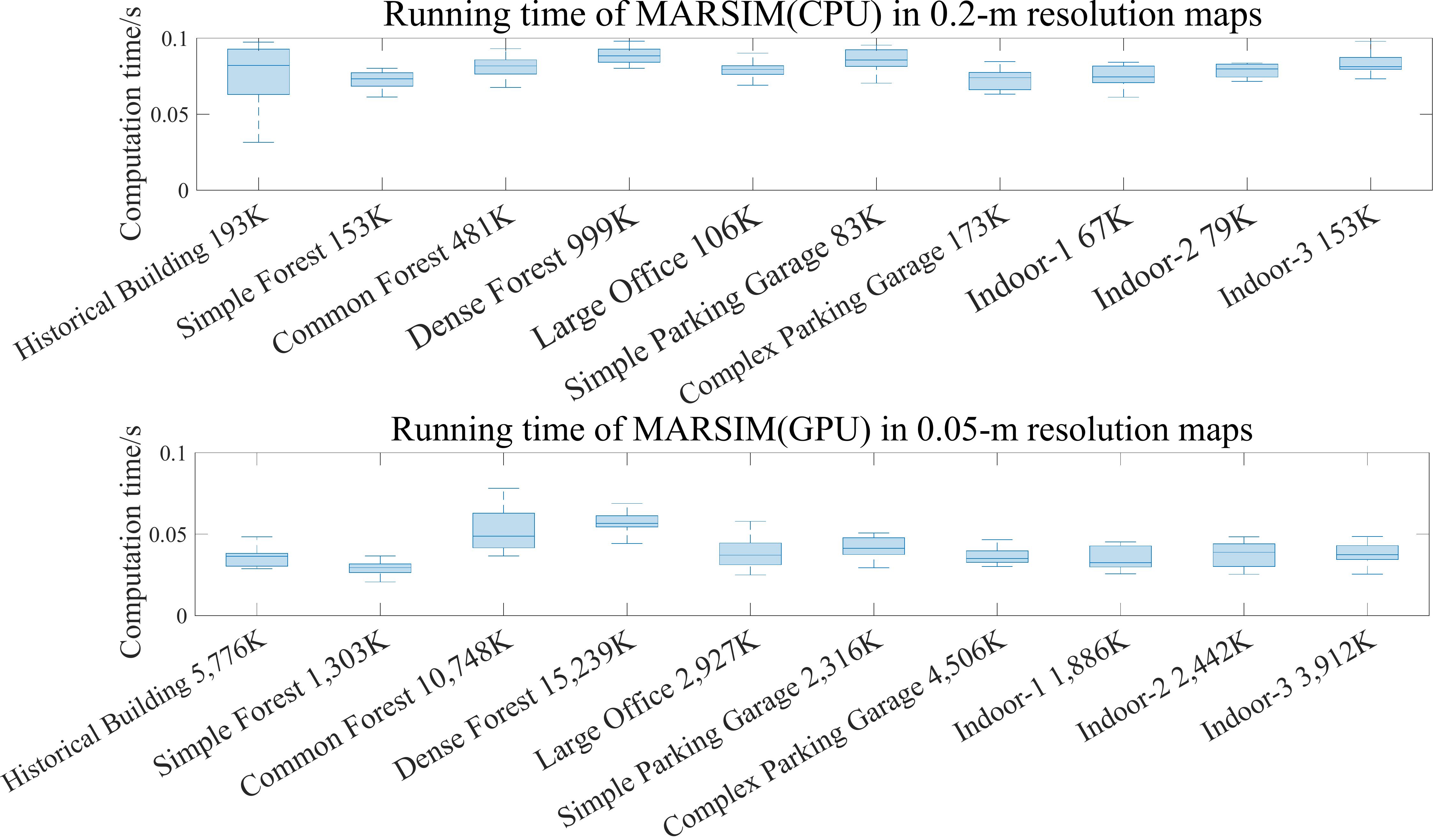}
		\caption{{Time consumption for rendering one Livox AVIA scan on all 10 realistic point cloud maps. The X-axis labels the name and size of the point cloud map.}}
		\label{Compute_time_analysis}
		\vspace{-5mm}
	\end{figure}

\subsection{Breakdown of computation resources consumption}   

We compare the time consumption between MARSIM simulator and the Gazebo simulator. Since Gazebo can only use mesh models, we transform the point cloud maps {of respective resolutions (see below)} to mesh models using Poisson reconstruction method\cite{poisson}. 
We select five typical scenes and compare the time and memory consumption of rendering one  scan of a Livox Avia LiDAR (77°$\times$70° FoV, 385$\times$350 resolution, 30-m sensing range), respectively. 
For each map, we test two cases: a high-resolution map (0.05m resolution) and a low-resolution map (0.2 resolution). The data is generated by randomly selecting 10 positions and yaw angles of the UAV. The running time comparison on a light-weight computing platform NUC 10 Kit (with an i7-10710U max frequency 4.70-GHz CPU, 32-G RAM) is shown in Fig. \ref{Gazebocomparison}. {It can be seen that in low-resolution maps, even the CPU version of MARSIM can achieve slightly less computation time than the GPU-accelerated Gazebo simulation. With GPU acceleration, MARSIM is two times faster than Gazebo. In high-resolution maps, the difference is even more obvious: the CPU version of MARSIM is two times faster than the GPU-accelerated Gazebo simulation while the GPU version of MARSIM is ten times faster.
The reason why Gazebo performed poorly in the experiments is because of the large number (over 2 million) of triangular faces in the generated mesh maps, which is necessary to retain a level of detail similar to the corresponding point cloud. In contrast, most existing robot simulations use very simple mesh maps, which are mainly composed of large planes and have a small number of triangular faces, which can be simulated in real time. Moreover, as a simulator specifically designed for point cloud, MARSIM does not need to process the whole render pipeline to render meshes (e.g., reducing the process of fragment shader, ray tracing, etc.) and complex physics simulation (like collision simulation), which decrease the consumption of computation resources significantly. 
}
Finally, we perform time consumption for all ten maps in 0.05-m and 0.2-m resolution with the same sensor. As shown in Fig. \ref{Compute_time_analysis}, in all cases, the simulator is able to run in real time at 10 Hz.

\begin{table}[]
\caption{{Memory consumption comparison with Gazebo on a light-weight computation platform (NUC).}}
\resizebox{0.48\textwidth}{25mm}{
\begin{tabular}{clcll}
\hline
\multirow{3}{*}{Resolution} & \multicolumn{1}{c}{\multirow{3}{*}{Map}} & \multicolumn{3}{c}{RAM Consumption (GB)}                                                        \\ \cline{3-5} 
                                & \multicolumn{1}{c}{}                     & \multicolumn{1}{l}{\multirow{2}{*}{Gazebo}} & \multicolumn{2}{c}{MARSIM}                        \\ \cline{4-5} 
                                & \multicolumn{1}{c}{}                     & \multicolumn{1}{l}{}                        & \multicolumn{1}{c}{CPU} & \multicolumn{1}{c}{GPU} \\ \hline
\multirow{5}{*}{0.2 m}            & Historical building                      & 1.64                                        & 1.01                    & 1.31                    \\
                                & Complex Parking Garage                             & 1.46                                        & 1.1                     & 1.25                    \\
                                & Large Office                             & 1.68                                        & 1.04                    & 1.13                    \\
                                & Common Forest                            & 1.48                                        & 1.37                    & 1.73                    \\
                                & Indoor-2                                 & 2.09                                        & 0.94                    & 1.14                    \\ \hline
\multirow{5}{*}{0.05 m}           & Historical building                      & 6.97                                        & 4.04                    & 3.43                    \\
                                & Complex Parking Garage                             & 6.72                                        & 3.51                    & 3.06                    \\
                                & Large Office                             & 4.93                                        & 2.87                    & 2.17                    \\
                                & Common Forest                            & 16.88                                       & 7.35                    & 3.53                    \\
                                & Indoor-2                                 & 3.73                                        & 2.51                    & 1.84                    \\ \hline
\end{tabular}}
\label{tab:RAMcomsumption}
\vspace{-20mm}
\end{table}

\begin{figure}[t]
	\centering
	\includegraphics[width=0.49\textwidth]{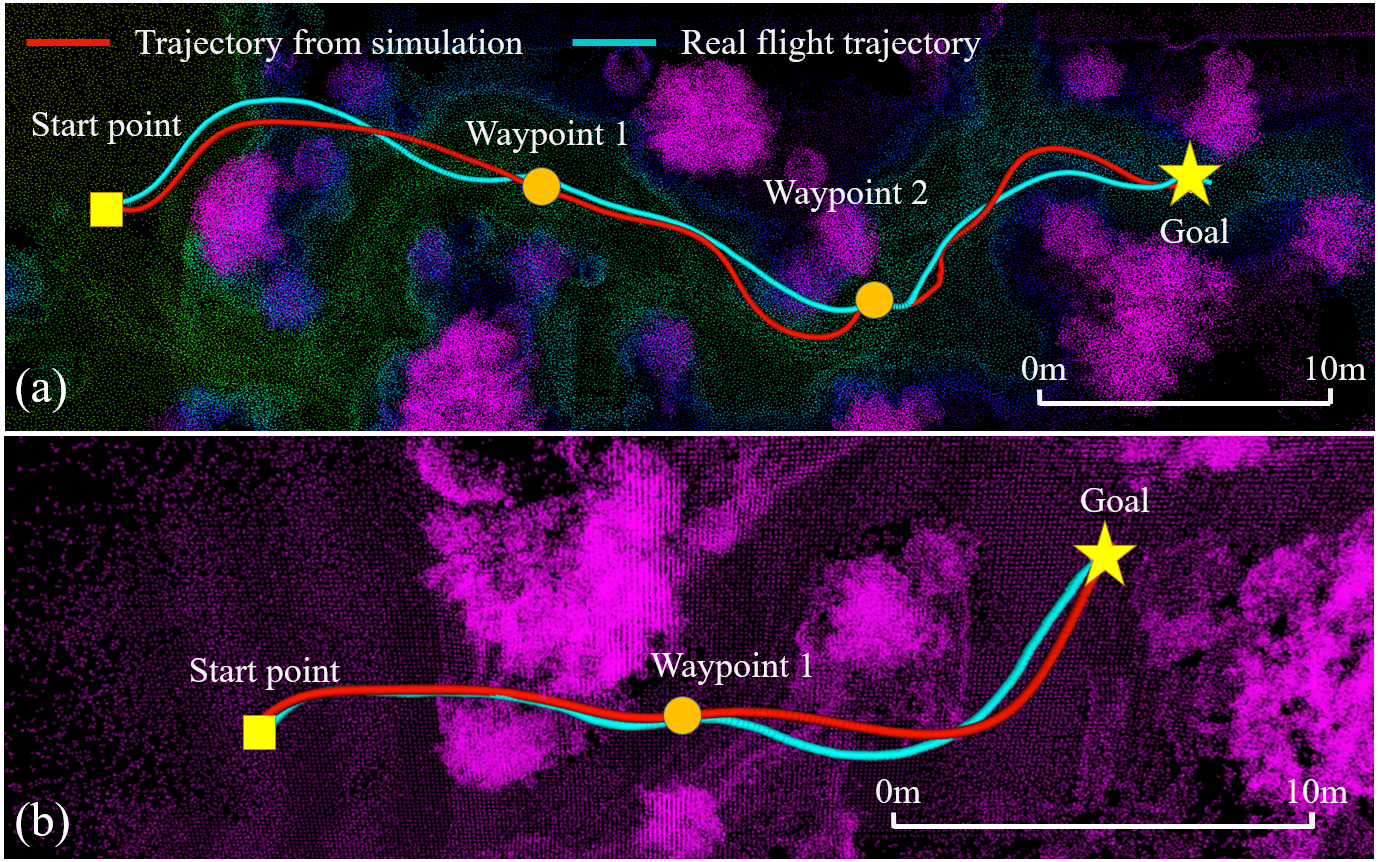}
	\caption{Comparisons between flight trajectories in simulation and actual environment. (a) and (b) represent two experiments in different environments. The red lines are the trajectory results from the simulation, the blue lines are the actual flight trajectories.  }
	\label{planning_comp}
\end{figure}

\begin{figure}[t]
	\centering
	\includegraphics[width=0.49\textwidth]{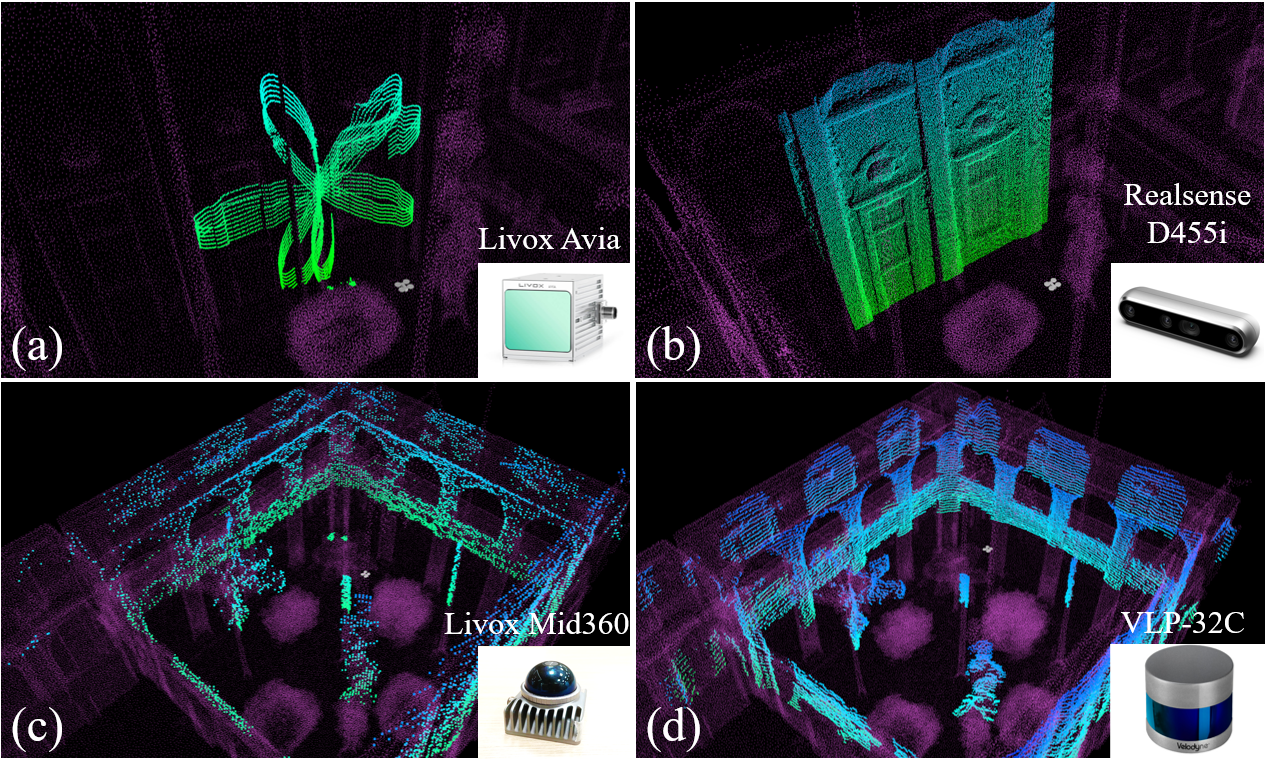}
	\caption{Various LiDAR scan pattern support including Livox Avia (a), D455 (b), Livox Mid-360 (c), and VLP 32 (d), respectively.}
	\label{fig:Multilidarsupport}
 \vspace{-18mm}
\end{figure}

In addition to the time consumption comparison, we also collected the RAM consumption as shown in Table \ref{tab:RAMcomsumption}. The RAM consumption of our simulator is about half that of the Gazebo simulator in both the CPU and GPU versions, which also demonstrates the light-weight characteristics of our simulator.

\subsection{Experiment verification}

To demonstrate that our simulator can provide flight simulation similar to real experiments, we verify the simulation of a UAV planning method, the Bubble planner \cite{bubbleplanner}, in an actual environment. To do so, we build a quadrotor UAV equipped with a Livox Mid360 LiDAR as used in \cite{bubbleplanner}. Then, we handheld the UAV to manually scan the actual environment, which is a forest scene, and build its point cloud map. We simulate the Bubble planner in our simulator using the collected point cloud map and compare the simulated UAV flight trajectory with that of the real experiment with the same start position and target position.  The comparison is shown in Fig. \ref{planning_comp}. As can be seen, the trajectories from the simulation are very close to the actual ones, which verifies the practicability of this simulator. 

\vspace{-5mm}

\begin{figure}[t]
	\centering
	\includegraphics[width=0.49\textwidth]{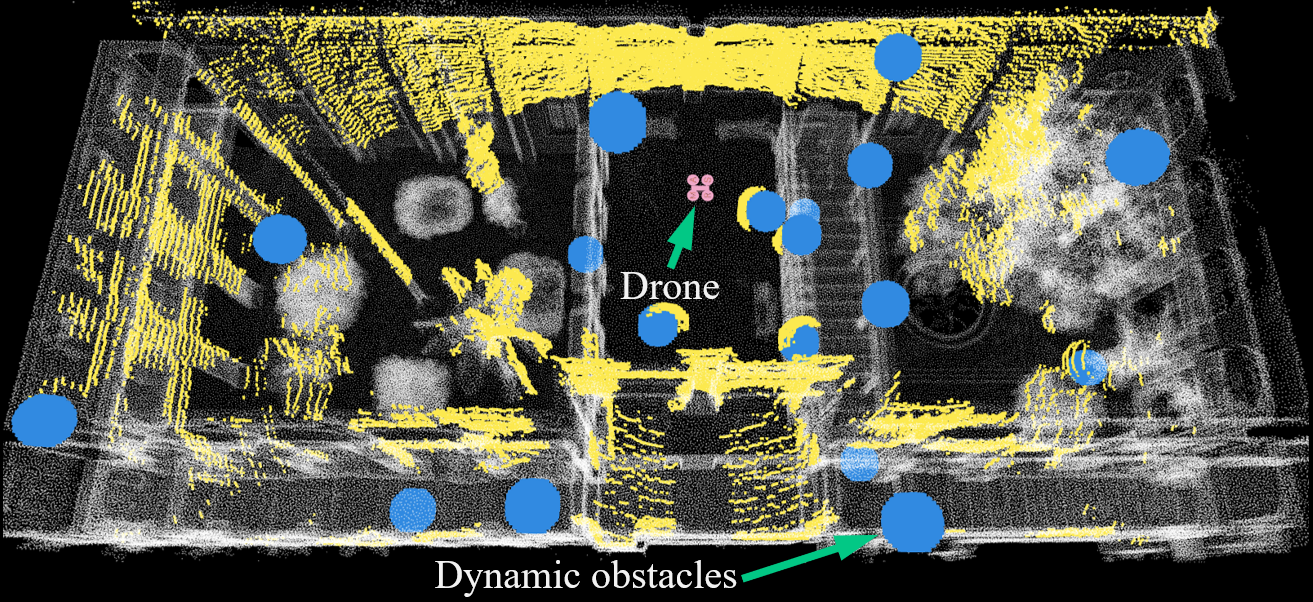}
	\caption{Dynamic obstacles simulation in the historical building map. The blue balls are the dynamic obstacles, the yellow point clouds are the points scanned by a VLP-32 LiDAR.}
	\label{fig:DynamicObs}
\end{figure}

\begin{figure}[t]
	\centering
	\includegraphics[width=0.49\textwidth]{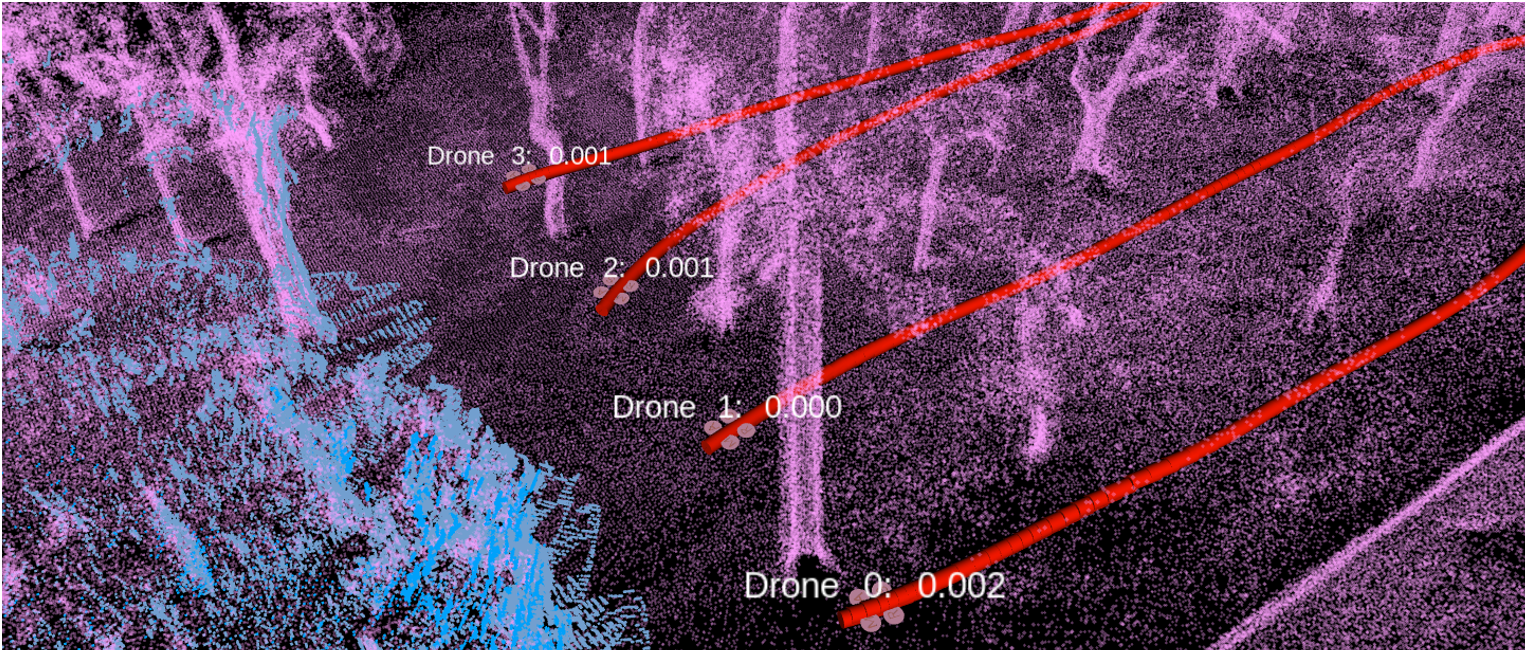}
	\caption{Multi-UAV planning simulation. The pink models are the UAVs, and the red curves are the trajectories of the drones, avoiding the obstacles of a realistic forest map.}
	\label{fig:Multiuav}
\end{figure}

\begin{figure}[t!]
	\centering
	\includegraphics[width=0.49\textwidth]{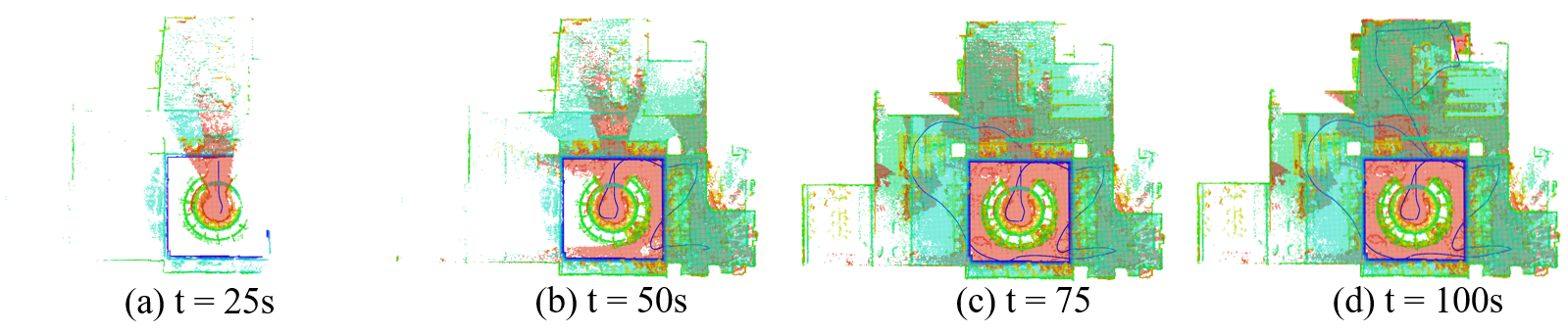}
	\caption{Demonstration of a UAV autonomous exploration simulation in indoor-2 map, utilizing FUEL algorithm. The area scanned by the UAV after different executing times are shown.}
	\label{explorationdemo}
	\vspace{-5mm}
\end{figure}

\subsection{{Support of different types of LiDARs and other functions}}
In order to increase the simulator's versatility, a variety of common LiDAR and depth camera models are also provided in the simulator. As shown in Figure \ref{fig:Multilidarsupport}, the simulator supports sensors such as Livox Avia, Livox Mid-360, VLP-32, VLP-64, OS1-32, and Intel realsense D455. The simulator can reproduce the scanning patterns of these sensors so that users can use them directly without tuning any parameters. 
{Moreover, dynamic obstacles and multi-UAV simulations are also supported, as shown in Fig. \ref{fig:DynamicObs} and Fig. \ref{fig:Multiuav}.}

\subsection{Practical applications of the simulator}

This simulator is mainly used to provide a testing and verification platform for the algorithm development of LiDAR-based UAVs, especially motion planning and autonomous exploration algorithms that require interaction with the environments. While previous experiments have shown the application of our simulator in UAV motion planning, we also carried out simulation experiments of autonomous UAV exploration. Fig. \ref{explorationdemo} shows the autonomous exploration process of a UAV carrying Livox Avia using FUEL\cite{zhou_fuel} algorithm in the indoor-2 map. 
It is worth mentioning that the simulator has been successfully used to assist the development of multi-UAV mutual location in \cite{fangchengUAVs} and motion planning algorithm in \cite{bubbleplanner,renyunfanSE3}.


\section{Conclusion and Discussion}
This paper proposes a LiDAR-based UAV simulator for real environment simulation on light-weight computing platforms. The simulator renders LiDAR scans directly on point cloud maps, which is way easier to capture for real environments than mesh models used by existing simulators. Moreover, due to the high accuracy of modern 3D LiDARs and laser scanners, a point cloud map scanned from real environments can truthfully represent the environment, which dramatically bridges the gap between simulation and reality. To maximize the practicality of the simulator, we further provide ten high-resolution point cloud maps and support the simulation of various types of LiDAR sensors, dynamic obstacles, and multi-UAV simulation. These features can meet the research and development needs of motion planning algorithms and autonomous exploration algorithms of single or multiple UAVs.

Since the simulator is based on point cloud maps, when the accuracy of the map is not high enough or there are noise points, the simulator cannot restore the correct details of the real environments. Also, the advantage in the computation efficiency of the simulator may degrade if a LiDAR scan is sparse, where the time for ray-casting on mesh models used by Gazebo (and other existing simulators) is reduced significantly. In contrast, the rendering module of our simulator has to generate a dense depth image not just on the scanning pattern, which leads to a waste of computing resources to calculate unnecessary depth image pixels. It could be an improvement direction of the simulator in the future.

\bibliography{IEEEabrv,Bibliography}

\begin{thebibliography}{10}
\providecommand{\url}[1]{#1}
\csname url@rmstyle\endcsname
\providecommand{\newblock}{\relax}
\providecommand{\bibinfo}[2]{#2}
\providecommand\BIBentrySTDinterwordspacing{\spaceskip=0pt\relax}
\providecommand\BIBentryALTinterwordstretchfactor{4}
\providecommand\BIBentryALTinterwordspacing{\spaceskip=\fontdimen2\font plus
\BIBentryALTinterwordstretchfactor\fontdimen3\font minus
  \fontdimen4\font\relax}
\providecommand\BIBforeignlanguage[2]{{%
\expandafter\ifx\csname l@#1\endcsname\relax
\typeout{** WARNING: IEEEtran.bst: No hyphenation pattern has been}%
\typeout{** loaded for the language `#1'. Using the pattern for}%
\typeout{** the default language instead.}%
\else
\language=\csname l@#1\endcsname
\fi
#2}}

\bibitem{dang_graphbased_2020}
T.~Dang, M.~Tranzatto, S.~Khattak, F.~Mascarich, K.~Alexis, and M.~Hutter,
  ``\BIBforeignlanguage{en}{Graph‐based subterranean exploration path
  planning using aerial and legged robots},''
  \emph{\BIBforeignlanguage{en}{Journal of Field Robotics}}, vol.~37, no.~8,
  pp. 1363--1388, Dec. 2020.

\bibitem{shah2020multidrone}
K.~Shah, G.~Ballard, A.~Schmidt, and M.~Schwager, ``Multidrone aerial surveys
  of penguin colonies in antarctica,'' \emph{Science Robotics}, vol.~5, no.~47,
  p. eabc3000, 2020.

\bibitem{vacanas2015building}
Y.~Vacanas, K.~Themistocleous, A.~Agapiou, and D.~Hadjimitsis, ``Building
  information modelling (bim) and unmanned aerial vehicle (uav) technologies in
  infrastructure construction project management and delay and disruption
  analysis,'' in \emph{Third International Conference on Remote Sensing and
  Geoinformation of the Environment (RSCy2015)}, vol. 9535.\hskip 1em plus
  0.5em minus 0.4em\relax SPIE, 2015, pp. 93--103.

\bibitem{bubbleplanner}
Y.~Ren, F.~Zhu, W.~Liu, Z.~Wang, Y.~Lin, F.~Gao, and F.~Zhang, ``Bubble
  planner: Planning high-speed smooth quadrotor trajectories using receding
  corridors,'' \emph{arXiv preprint arXiv:2202.12177 (accepted by 2022 IROS)},
  2022.

\bibitem{gazebo}
N.~Koenig and A.~Howard, ``Design and use paradigms for gazebo, an open-source
  multi-robot simulator,'' in \emph{2004 IEEE/RSJ International Conference on
  Intelligent Robots and Systems (IROS) (IEEE Cat. No.04CH37566)}, vol.~3,
  2004, pp. 2149--2154 vol.3.

\bibitem{webots}
O.~Michel, ``Cyberbotics {Ltd}. {Webots}™: {Professional} {Mobile} {Robot}
  {Simulation},'' \emph{International Journal of Advanced Robotic Systems},
  vol.~1, no.~1, p.~5, Mar. 2004, publisher: SAGE Publications.

\bibitem{shah2018airsim}
S.~Shah, D.~Dey, C.~Lovett, and A.~Kapoor, ``Airsim: High-fidelity visual and
  physical simulation for autonomous vehicles,'' in \emph{Field and service
  robotics}.\hskip 1em plus 0.5em minus 0.4em\relax Springer, 2018, pp.
  621--635.

\bibitem{poisson}
M.~Kazhdan, M.~Bolitho, and H.~Hoppe, ``Poisson surface reconstruction,'' in
  \emph{Proceedings of the fourth Eurographics symposium on Geometry
  processing}, vol.~7, 2006.

\bibitem{matlab_quadrotor}
F.~{\c{S}}enkul and E.~Altu{\u{g}}, ``Modeling and control of a novel
  tilt—roll rotor quadrotor uav,'' in \emph{2013 International Conference on
  Unmanned Aircraft Systems (ICUAS)}.\hskip 1em plus 0.5em minus 0.4em\relax
  IEEE, 2013, pp. 1071--1076.

\bibitem{lvximin_tailsitter}
L.~Ximin, H.~Gu, J.~Zhou, Z.~Li, S.~Shen, and F.~Zhang, ``Simulation and flight
  experiments of a quadrotor tail-sitter vertical take-off and landing unmanned
  aerial vehicle with wide flight envelope,'' \emph{International Journal of
  Micro Air Vehicles}, vol.~10, pp. 303--317, 12 2018.

\bibitem{gazeboplanning}
W.~Luo, Q.~Tang, C.~Fu, and P.~Eberhard, ``Deep-sarsa based multi-uav path
  planning and obstacle avoidance in a dynamic environment,'' in
  \emph{International Conference on Swarm Intelligence}.\hskip 1em plus 0.5em
  minus 0.4em\relax Springer, 2018, pp. 102--111.

\bibitem{airsim_gaofei}
Z.~Han, Z.~Wang, N.~Pan, Y.~Lin, C.~Xu, and F.~Gao, ``Fast-racing: An
  open-source strong baseline for $\mathrm{SE}(3)$ planning in autonomous drone
  racing,'' \emph{IEEE Robotics and Automation Letters}, vol.~6, no.~4, pp.
  8631--8638, 2021.

\bibitem{rong_lgsvl_2020}
G.~Rong, B.~H. Shin, H.~Tabatabaee, Q.~Lu, S.~Lemke, M.~Možeiko, E.~Boise,
  G.~Uhm, M.~Gerow, S.~Mehta, E.~Agafonov, T.~H. Kim, E.~Sterner, K.~Ushiroda,
  M.~Reyes, D.~Zelenkovsky, and S.~Kim, ``{LGSVL} {Simulator}: {A} {High}
  {Fidelity} {Simulator} for {Autonomous} {Driving},'' \emph{arXiv:2005.03778
  [cs, eess]}, June 2020, arXiv: 2005.03778.

\bibitem{mbplanner}
M.~Dharmadhikari, T.~Dang, L.~Solanka, J.~Loje, H.~Nguyen, N.~Khedekar, and
  K.~Alexis, ``Motion primitives-based path planning for fast and agile
  exploration using aerial robots,'' 05 2020, pp. 179--185.

\bibitem{cao_tare_2021}
C.~Cao, H.~Zhu, H.~Choset, and J.~Zhang, ``\BIBforeignlanguage{en}{{TARE}: {A}
  {Hierarchical} {Framework} for {Efficiently} {Exploring} {Complex} {3D}
  {Environments}},'' in \emph{\BIBforeignlanguage{en}{Robotics: {Science} and
  {Systems} {XVII}}}.\hskip 1em plus 0.5em minus 0.4em\relax Robotics: Science
  and Systems Foundation, July 2021.

\bibitem{splatplanner}
A.~Brunel, A.~Bourki, C.~Demonceaux, and O.~Strauss, ``Splatplanner: Efficient
  autonomous exploration via permutohedral frontier filtering,'' in \emph{IEEE
  International Conference on Robotics and Automation (ICRA 2021)}, 2021.

\bibitem{brunel_flybo_2021}
A.~Brunel, A.~Bourki, O.~Strauss, and C.~Demonceaux,
  ``\BIBforeignlanguage{en}{{FLYBO}: {A} {Unified} {Benchmark} {Environment}
  for {Autonomous} {Flying} {Robots}},'' in \emph{\BIBforeignlanguage{en}{2021
  {International} {Conference} on {3D} {Vision} ({3DV})}}.\hskip 1em plus 0.5em
  minus 0.4em\relax London, United Kingdom: IEEE, Dec. 2021, pp. 1420--1431.

\bibitem{darpa}
\BIBentryALTinterwordspacing
N.~Koenig, ``Darpa subt virtual competition software,'' 2019. [Online].
  Available: \url{https://github.com/osrf/subt/tree/master/subt_ign/worlds}
\BIBentrySTDinterwordspacing

\bibitem{Lidarsim}
S.~Manivasagam, S.~Wang, K.~Wong, W.~Zeng, M.~Sazanovich, S.~Tan, B.~Yang,
  W.-C. Ma, and R.~Urtasun, ``Lidarsim: Realistic lidar simulation by
  leveraging the real world,'' in \emph{Proceedings of the IEEE/CVF Conference
  on Computer Vision and Pattern Recognition}, 2020, pp. 11\,167--11\,176.

\bibitem{surfels}
H.~Pfister, M.~Zwicker, J.~Baar, and M.~Gross, ``Surfels: Surface elements as
  rendering primitives,'' \emph{Proceedings of the ACM SIGGRAPH Conference on
  Computer Graphics}, 05 2000.

\bibitem{tsdf}
A.~Millane, Z.~Taylor, H.~Oleynikova, J.~Nieto, R.~Siegwart, and C.~Cadena,
  ``C-blox: A scalable and consistent tsdf-based dense mapping approach,'' in
  \emph{2018 IEEE/RSJ International Conference on Intelligent Robots and
  Systems (IROS)}, 2018.

\bibitem{delauncytriangulation}
P.~Labatut, J.-P. Pons, and R.~Keriven, ``Efficient multi-view reconstruction
  of large-scale scenes using interest points, delaunay triangulation and graph
  cuts,'' in \emph{2007 IEEE 11th International Conference on Computer Vision},
  2007, pp. 1--8.

\bibitem{r3live}
J.~Lin and F.~Zhang, ``R3live: A robust, real-time, rgb-colored,
  lidar-inertial-visual tightly-coupled state estimation and mapping package,''
  in \emph{2022 International Conference on Robotics and Automation (ICRA)},
  2022, pp. 10\,672--10\,678.

\bibitem{flightgoggles}
W.~Guerra, E.~Tal, V.~Murali, G.~Ryou, and S.~Karaman, ``Flightgoggles:
  Photorealistic sensor simulation for perception-driven robotics using
  photogrammetry and virtual reality,'' in \emph{2019 IEEE/RSJ International
  Conference on Intelligent Robots and Systems (IROS)}, 2019, pp. 6941--6948.

\bibitem{fastlio2}
W.~Xu, Y.~Cai, D.~He, J.~Lin, and F.~Zhang, ``Fast-lio2: Fast direct
  lidar-inertial odometry,'' \emph{IEEE Transactions on Robotics}, pp. 1--21,
  2022.

\bibitem{xiyuanBA}
X.~Liu, Z.~Liu, F.~Kong, and F.~Zhang, ``Large-scale lidar consistent mapping
  using hierachical lidar bundle adjustment,'' \emph{arXiv preprint
  arXiv:2209.11939}, 2022.

\bibitem{zhou_fuel}
B.~Zhou, Y.~Zhang, X.~Chen, and S.~Shen, ``Fuel: Fast uav exploration using
  incremental frontier structure and hierarchical planning,'' \emph{IEEE
  Robotics and Automation Letters}, vol.~PP, pp. 1--1, 01 2021.

\bibitem{opengl}
D.~Shreiner, B.~T. K. O. A.~W. Group, \emph{et~al.}, \emph{OpenGL programming
  guide: the official guide to learning OpenGL, versions 3.0 and 3.1}.\hskip
  1em plus 0.5em minus 0.4em\relax Pearson Education, 2009.

\bibitem{dynamicmodel}
T.~Lee, M.~Leok, and N.~H. McClamroch, ``Geometric tracking control of a
  quadrotor uav on se(3),'' in \emph{49th IEEE Conference on Decision and
  Control (CDC)}, 2010, pp. 5420--5425.

\bibitem{propellermodel}
R.~Mahony, V.~Kumar, and P.~Corke, ``Multirotor aerial vehicles: Modeling,
  estimation, and control of quadrotor,'' \emph{IEEE Robotics \& Automation
  Magazine}, vol.~19, no.~3, pp. 20--32, 2012.

\bibitem{moderncontrol}
K.~Ogata \emph{et~al.}, \emph{Modern control engineering}.\hskip 1em plus 0.5em
  minus 0.4em\relax Prentice hall Upper Saddle River, NJ, 2010, vol.~5.

\bibitem{fastlivo}
C.~Zheng, Q.~Zhu, W.~Xu, X.~Liu, Q.~Guo, and F.~Zhang, ``Fast-livo: Fast and
  tightly-coupled sparse-direct lidar-inertial-visual odometry,'' \emph{ArXiv},
  vol. abs/2203.00893, 2022.

\bibitem{fangchengUAVs}
F.~Zhu, Y.~Ren, F.~Kong, H.~Wu, S.~Liang, N.~Chen, W.~Xu, and F.~Zhang,
  ``Decentralized lidar-inertial swarm odometry,'' \emph{arXiv preprint
  arXiv:2209.06628}, 2022.

\bibitem{renyunfanSE3}
Y.~Ren, S.~Liang, F.~Zhu, G.~Lu, and F.~Zhang, ``Online whole-body motion
  planning for quadrotor using multi-resolution search,'' \emph{arXiv preprint
  arXiv:2209.06761}, 2022.

\end{thebibliography}

\vfill

\end{document}